\pdfoutput=1

\documentclass[11pt]{article}
\usepackage{algorithmicx}
\usepackage{algpseudocode}
\usepackage{amsmath} 
\usepackage{graphicx}
\usepackage{amsfonts,amssymb}
\usepackage{amsmath}
\usepackage{multirow}
\usepackage{booktabs}
\usepackage{algorithm}
\usepackage{tabularx} 
\usepackage{booktabs} 

\DeclareMathOperator*{\argmax}{arg\,max}

\usepackage{ACL2023}
\usepackage{times}
\usepackage{latexsym}

\usepackage[T1]{fontenc}

\usepackage[utf8]{inputenc}

\usepackage{microtype}

\usepackage{inconsolata}

\title{Latent Distance Guided Alignment Training for Large Language Models}

\author{Haotian Luo \\
  Sichuan University \\
}

\begin{document}
\maketitle
\begin{abstract}

Ensuring alignment with human preferences is a crucial characteristic of large language models (LLMs). Presently, the primary alignment methods, RLHF and DPO, require extensive human annotation, which is expensive despite their efficacy. The significant expenses associated with current alignment techniques motivate researchers to investigate the development of annotation-free alignment training methods. In pursuit of improved alignment without relying on external annotation, we introduce Latent Distance Guided Alignment Training (LD-Align). This approach seeks to align the model with a high-quality supervised fine-tune dataset using guidance from a latent space. The latent space is generated through sample reconstruction, akin to auto-encoding. Consequently, we utilize the distance between sample pairs in the latent space to guide DPO-based alignment training. Extensive experimentation and evaluation show the efficacy of our proposed method in achieving notable alignment.\footnote{The V1 unfinished version of the submission has never been reviewed and approved by Huaxiu Yao. We will release a finalized version soon.}
\end{abstract}
\section{Introduction}
Over the past two years, LLMs have demonstrated strong performance. LLMs have shown remarkable performance in various fields of NLP, such as mathematical problem solving, summarization generation, reading comprehension, and open-ended question answering, achieving notable results. In order to align the behavior of LLMs with human expectations, such as adhering to facts and avoiding biases, and to better elicit their capabilities, such as mathematical reasoning, researchers have proposed alignment training methods, which typically involve a process requiring extensive manual annotation of data. Aligment training is typically employed after Supervised Fine-tune(SFT), with the most commonly used mainstream methods being Reinforcement Learning with Human Feedback (RLHF) and Direct Preference Optimization (DPO)\cite{NEURIPS2023_a85b405e}. 

Since mainstream alignment training methods typically require extensive manual annotation, which is expensive to obtain, the pursuit of an alignment method that does not necessitate human annotation is becoming increasingly popular. To solve this challenging problem, there are currently some efforts aimed at avoiding manual annotation in alignment tasks. RLAIF\cite{lee2023rlaif} utilizes large language models to generate preference labels instead of human annotators and explores the direct utilization of language models to generate reward scores. SPIN\cite{chen2024self} iteratively train a LLM to align on the SFT datasets through a self-play mechanism which shares a similar motivation with GAN\cite{goodfellow2020generative}. Also, \cite{yuan2024self} have studied Self-Rewarding Language Models, where the language model itself is used via LLM-as-a-Judge prompting to provide its own rewards during training.


In the present study, we introduce a DPO-based novel approach termed LD-Align (Latent Distance Guided Alignment Training), aimed at iteratively aligning a fine-tuned LLM with a given high-quality SFT dataset without any additional human annotation or reliance on a more powerful LLM for support. Within this framework, we consider samples sourced from the SFT dataset as golden labels, contrasting with those generated by the model, which we categorize as dispreferred samples. By quantifying the disparity between the latent feature vectors of these two sets within a latent space established through sample reconstruction, we ascertain the alignment status of each instance, effectively gauging the degree of suboptimality. Subsequent to this assessment, employing iterative training facilitated by DPO, we assign higher update weights to samples exhibiting lower alignment levels, thus stimulating exploration for potential better alignment, while conversely allotting smaller update weights to those with higher alignment levels, thereby mitigating the risk of overfitting. Our contributions can be succinctly summarized as follows:

\begin{itemize}
    \item We propose a novel DPO-based align method (LD-Align), which is guided in instance-level in the training process.
    \item Through comprehensive experiments, the proposed method achieves the best performances among selected competing methods.
\end{itemize}

\section{Related Work}
\subsection{Aligning LLM with human preference}
With advancements in AI systems, the accompanying rise in risks becomes more pronounced. Undesirable behaviors exhibited by LLMs, such as providing untruthful responses, displaying sycophancy, and engaging in deception, become exacerbated as the scale of the models increases. This phenomenon raises concerns regarding the management of advanced AI systems, which are increasingly challenging to regulate. The predominant alignment approach commonly employed involves learning from feedback. A conventional strategy is reinforcement learning from human feedback (RLHF). Here, human assessors furnish feedback by comparing various responses generated by the chat model, and this feedback is subsequently utilized through Reinforcement Learning (RL) in conjunction with a pre-trained reward model, which is often trained with PPO. Due to the complexity of the process and the instability encountered during training, researchers have introduced Direct Performance Optimization. This has been demonstrated by the community to be an effective approach.



\section{Preliminaries}
We consider a Large Language Model (LLM) parameterized by $\mathbf{\theta}$ and denoted as $p_{\mathbf{\theta}}$, which accepts a sequence $\mathbf{x} = [x_1, \ldots, x_n]$, commonly termed as the prompt, and then generate a corresponding response $\mathbf{y} = [y_1, \ldots, y_m]$. Hence, the response $\mathbf{y}$ is construed as a sample drawn from the conditional probability distribution $p_{\mathbf{\theta}}(\cdot | \mathbf{x})$. The conditional probability distribution $p_{\mathbf{\theta}}(\mathbf{y} | \mathbf{x})$ can be decomposed as follows:
\begin{align}
p_{\mathbf{\theta}}(\mathbf{y} | \mathbf{x}) = \prod_{j=1}^{m} p_{\mathbf{\theta}}(y_{j} | \mathbf{x}, \mathbf{y}_{<j}),
\end{align}
Subsequently, we review supervised fine-tuning (SFT), which is the primary training methodologies to adapt a pre-trained LLM for downstream tasks, utilizing a relatively smaller dataset of labeled examples compared to the data used in pre-training stage.
In this paper, we focus on the task of instruction-tuning where the prompt-answer pairs denoted as $(\mathbf{x},\mathbf{y})$, are drawn from a specified SFT dataset $\mathcal{D}$. Thus the training objective of SFT under the instruction tuning setting can be formulated as:
\begin{align*}
\max_{p_{\theta}}  \mathbb{E}_{(x,y)\sim \mathcal{D}} 
 \Big[\log p_{\mathbf{\theta}}(\mathbf{y} | \mathbf{x})\Big]
\end{align*}

\section{Method}
In this section, we introduce our new fine-tuning method (Self-Aligning in Latent Space) for boosting LLMs given a high-quality supervised fine-tuning (SFT) dataset $\mathcal{D} = \{(\mathbf{x}, \mathbf{y})\}_{N}$ containing N samples, without relying on additional external annotations. In summary, given a LLM $p_{\mathbf{\theta}}$ fine-tuned on $\mathcal{D}$, we train a guiding model through auto-encoding to assess the distance between generated samples and real samples within the latent space of this model. Subsequently, we employ this distance as a signal to guide the model aligning with the real data distribution via direct preference optimization (DPO).
\subsection{Guiding model}
Considering an auto-encoding structure $\mathcal{T} = (\mathbf{\phi},\mathbf{\psi})$ with conditions, which consists of an encoder $\mathbf{\phi}$ and a decoder $\mathbf{\psi}$. Both encoder and decoder are based on pre-trained transformers, which are set to be GPT-2\cite{radford2019language} in our experiment. The encoder $\mathbf{\phi}$ takes prompt $\mathbf{x}$ and response $\mathbf{y}$ as input, and output a multi-dimension latent vector $\mathbf{h}=\mathbf{\phi}(\mathbf{x},\mathbf{y}) \in \mathbb{R}^{d}$. Meanwhile, decoder $\mathbf{\psi}$ takes prompt $\mathbf{x}$ and latent variable $\mathbf{h}$ as input and try to reconstruct the real response $\mathbf{y}$. The training data for the model is sourced from two components: the SFT dataset $\mathcal{D}$ and generated samples $\{(x, y^{\prime})\}_{i=1}^{n}$ produced by the initial fine-tuned model $p_{\mathbf{\theta}}$.
The training objective of guiding model $\mathcal{T}$ can be formulated as:

\begin{algorithm*}
\begin{align}
    \mathcal{L}&_{guide}(\mathbf{\psi},\mathbf{\phi})= - \mathbb{E}_{(\mathbf{x},\mathbf{y}) \sim \mathcal{D},y^{\prime}\sim p_{\theta}(\cdot \mid x)  } \Big[\log {p_\mathbf{\psi}}(\mathbf{y} | \mathbf{x}, \mathbf{\phi}(\mathbf{x},\mathbf{y}))+\log {p_\mathbf{\psi}}(\mathbf{y^\prime} | \mathbf{x}, \mathbf{\phi}(\mathbf{x},\mathbf{y^\prime}))\Big]
\end{align}

\begin{align}
\max_{p_{\theta}}  \mathbb{E}_{(x,y)\sim \mathcal{D}, y^{\prime}\sim p_{\theta}(\cdot \mid x)} \Big[\frac{\mathbf{s_{\phi}}(\mathbf{x},\mathbf{y},\mathbf{y^\prime})} {\mathbf{S_{\phi}}(\mathcal{D},p_{\theta})}\log p^*(\mathbf{y}\succ \mathbf{y^\prime} \mid x)\Big]
\end{align}

\begin{multline*}\tag{4}
\mathcal{L}_{\text{Align}}(p_{\theta}; p_{ref})= \\-\mathbb{E}_{(x,y)\sim \mathcal{D}, y^{\prime}\sim \pi_{\theta}(\cdot \mid x)}\left[\frac{\mathbf{s_{\phi}}(\mathbf{x},\mathbf{y},\mathbf{y^\prime})} {\mathbf{S_{\phi}}(\mathcal{D},p_{ref})} \log \sigma \left(\beta \log \frac{p_{\theta}(y\mid x)}{p_{ref}(y\mid x)} - \beta \log \frac{p_{\theta}(y^\prime\mid x)}{p_{ref}(y^\prime\mid x)}\right)\right]
\end{multline*}

\begin{multline*}\tag{5}
    \nabla_\theta \mathcal{L}_\text{Align}(p_\theta;p_{ref}) =\\  -\beta\mathbb{E}_{(x,y)\sim \mathcal{D}, y^{\prime}\sim \pi_{\theta}(\cdot \mid x)} \bigg[{\frac{\mathbf{s_{\phi}}(\mathbf{x},\mathbf{y},\mathbf{y^\prime})} {\mathbf{S_{\phi}}(\mathcal{D},p_{ref})}}\sigma(z)\bigg[\underbrace{\nabla_\theta\log p(y \mid x)}_\text{increase likelihood of $y$} - \underbrace{\nabla_\theta\log p(y^\prime \mid x)}_\text{decrease likelihood of $y^\prime$}\bigg]\bigg],
\end{multline*}

\end{algorithm*}
To minimize $\mathcal{L}_{guide}$, encoder $\mathbf{\phi}$ needs to extract some attributes in the form of a latent vector while decoder $\mathbf{\psi}$ needs to utilize this latent vector $\mathbf{\phi}(\mathbf{x},\mathbf{y})$ or $\mathbf{\phi}(\mathbf{x},\mathbf{y^\prime})$ to successfully reconstruct $\mathbf{y}$ or $\mathbf{y^\prime}$ given the same $\mathbf{x}$.
\subsection{Self-Aligning in Latent Space}
\subsubsection{Overview}
This subsection describes our main method deliberately. We regard $\mathbf{y}$ as the winner sample and $\mathbf{y^\prime}$ as the loser sample, and employ a DPO-based approach to train iteratively for alignment using the distance in the latent space as guidance. In each iteration, for samples with larger distances in the latent space, we assign greater update weights, while conversely, for samples with smaller distances in the latent space, indicating proximity to real samples, we assign smaller update weights.

\subsubsection{Details}
As $\mathcal{D}$ is a high-quality dataset, each $\mathbf{y}\in\mathcal{D}$ can be seen as a golden label. In other words, we will have a preference over a tuple ($\mathbf{y}$, $\mathbf{y^\prime}$) when given an $\mathbf{x}$, where $\mathbf{y}$ denotes a real response from SFT dataset while $\mathbf{y^\prime}$ denotes a response generated by $p_{\mathbf{\theta}}$. This preference can be denoted as $\mathbf{y}\succ \mathbf{y^\prime} \mid \mathbf{x}$. For each tuple ($\mathbf{y}$, $\mathbf{y^\prime}$) with a given $\mathbf{x}$, considering the setting of a Bradley-Terry model, we have:
\begin{align*}
    p^*(\mathbf{y}\succ \mathbf{y^\prime} \mid x)=\frac{\exp\left(r^*(\mathbf{x}, \mathbf{y})\right)}{\exp\left(r^*(\mathbf{x}, \mathbf{y})\right) + \exp\left(r^*(\mathbf{x}, \mathbf{y^\prime})\right)}.
\end{align*}
In this formula, $r^*$ is a  ground-truth reward model that cannot be accessed. With the guiding model $\mathcal{T} = (\mathbf{\phi},\mathbf{\psi})$ mentioned above, we can deriving the per-instance distance between two responses for a given prompt in latent space via:
\begin{align*}
\mathbf{s_{\phi}}(\mathbf{x},\mathbf{y},\mathbf{y^\prime}) = \left \| \mathbf{\phi}(\mathbf{x},\mathbf{y}) - \mathbf{\phi}(\mathbf{x},\mathbf{y^\prime}) \right \|_{2}
\end{align*}
 A large $\mathbf{s_{\phi}}$ indicates that the generated response is relatively far from the real response. Meanwhile, a smaller $\mathbf{s_{\phi}}$ means the generated response is already good enough as it is close to the real response in latent space. Now we can declare that, our goal is to align the language model $p_{\mathbf{\theta}}$ to the SFT dataset $\mathcal{D}$ with the guidance of normalized per-instance distances in latent space. Thus we can formulate an optimization problem.
where $\mathbf{S_{\phi}}(\mathcal{D},p_{\theta})$ is the expectation of distance on the SFT dataset $\mathcal{D}$ calculated by:
\begin{align*}
\mathbf{S_{\phi}}(\mathcal{D},p_{\theta}) = \frac{\mathbf{s_{\phi}}(\mathbf{x},\mathbf{y},\mathbf{y^\prime})} {\mathbb{E}_{(x,y)\sim \mathcal{D}, y^{\prime}\sim p_{\theta}(\cdot \mid x)}\Big[\mathbf{s_{\phi}}(\mathbf{x},\mathbf{y},\mathbf{y^\prime})\Big]}
\end{align*}
This optimization objective can be regarded as a weighted Bradley-Terry model, according to DPO, we can solve this optimization problem using the following approach:

\begin{algorithm*}
\caption{LD-Align Pseudo Code}
\begin{algorithmic}[1]
\Require SFT Dataset $\mathcal{D}=\{(\mathbf{x}_i, \mathbf{y}_i)\}_{i\in [N]}$, Guide model$\mathcal{T} = (\mathbf{\phi},\mathbf{\psi})$, LLM with parameter $\theta_{0}$
\For {$i = 1, 2, ...N$} 
\State Generate responses $\mathbf{y}_{i} \sim p_{\theta_{0}}(\cdot|\mathbf{x}_i)$.
\EndFor
\State Update $(\mathbf{\phi},\mathbf{\psi}) = \argmax_{(\mathbf{\phi},\mathbf{\psi})} \sum_{i\in [N]} {\log {p_\mathbf{\psi}}(\mathbf{y}_i | \mathbf{x}_i, \mathbf{\phi}(\mathbf{x}_i,\mathbf{y}_i))+\log {p_\mathbf{\psi}}(\mathbf{y^\prime}_i | \mathbf{x}_i, \mathbf{\phi}(\mathbf{x}_i,\mathbf{y^\prime}_i))}$

\For {$t = 1, 2, ...T$}
\For {$i = 1, 2, ...N$} 
\State Generate responses $\mathbf{y}_{i} \sim p_{\theta_{T-1}}(\cdot|\mathbf{x}_i)$.
\State Calculate distance $\mathbf{s_{\phi}}(\mathbf{x}_i,\mathbf{y}_i,\mathbf{y^\prime}_i) = \left \| \mathbf{\phi}(\mathbf{x}_i,\mathbf{y}_i) - \mathbf{\phi}(\mathbf{x}_i,\mathbf{y^\prime}_i) \right \|_{2}$
\EndFor
\State Calculate distance expectation $\mathbf{S_{\phi}}(\mathcal{D},p_{\theta_{T-1}}) = \frac{1}{N}\sum_{i\in [N]}{\mathbf{s_{\phi}}(\mathbf{x}_i,\mathbf{y}_i,\mathbf{y^\prime}_i)}$
\State Update   $\theta_{T} = \argmax_{\theta}\sum_{i\in [N]}{\frac{\mathbf{s_{\phi}}(\mathbf{x}_i,\mathbf{y}_i,\mathbf{y^\prime}_i)} {\mathbf{S_{\phi}}(\mathcal{D},p_{\theta_{T-1}})} \log \sigma \left(\beta \log \frac{p_{\theta}(\mathbf{y}_i\mid \mathbf{x}_i)}{p_{\theta_{T-1}}(\mathbf{y}_i\mid \mathbf{x}_i)} - \beta \log \frac{p_{\theta}(\mathbf{y^\prime}_i\mid \mathbf{x}_i)}{p_{\theta_{T-1}}(\mathbf{y^\prime}_i\mid \mathbf{x}_i)}\right)}$

\EndFor

\end{algorithmic}
\end{algorithm*}
\subsubsection{Evaluation}
For the optimization objective mentioned above, the gradient with respect to the parameters $\mathbf{\theta}$ can be written as:
where $\sigma(z)=\beta \log \frac{p_{\theta}(y\mid x)}{p_{ref}(y\mid x)} - \beta \log \frac{p_{\theta}(y^\prime\mid x)}{p_{ref}(y^\prime\mid x)}$. From this formulation, we can see that, for a pair of $y$ and $y^\prime$, we assign higher update weight when $y^\prime$ is far from $y$ in the latent space, which is denoted by the magnitude of normalized distance ${\frac{\mathbf{s_{\phi}}(\mathbf{x},\mathbf{y},\mathbf{y^\prime})} {\mathbf{S_{\phi}}(\mathcal{D},p_{ref})}}$. This re-weighting term is efficient because of that, in a standard preference optimization problem, we only ranked samples instead of golden labels while when given a golden dataset (for example, high-quality SFT dataset $\mathcal{D}$), we can measure how well is the LLM aligned with golden samples. The re-weighting term derived from reconstruction can allocate more attention to the samples that are not good enough and allocate less attention to those are already excellent to avoid over-fitting during the training procedure.

\section{Experiment}

\begin{table*}[h]
\centering
\resizebox{1.0\linewidth}{!}{
\begin{tabular}{lcccccccc}
\toprule
Model           & TruthfulQA & PIQA & HellaSwag & ARC & OpenbookQA & GSM8k & MuTual &  AVG\\
\midrule
Zephyr-sft      & 43.7 & 82.1 & 62.1 & 55.7 & 30.8 & 44.4 & 63.5  & -\\
SPIN            & 48.7 & 81.7 & 65.1 & 60.4 & 33.6 & 45.4 & 61.2  & 4.56\% \\
\midrule
LD-Align 0      & 49.5 & 82.2 & 65.0 & 59.1 & 33.2 & 44.3 & 63.3  & 4.49\% \\
LD-Align 1      & 50.7 & 82.4 & 66.1 & 59.7 & 33.6 & 44.9 & 63.5  & 5.70\%\\
LD-Align 2      & 50.3 & 82.5 & 66.2 & 60.7 & 33.2 & 45.9 & 63.3  & 6.00\%\\
\bottomrule
\end{tabular}}
\label{tab:my_table}
\end{table*}

\subsection{Experiment setup}
\subsubsection{Dataset}
The SFT dataset we use is Ultrachat200k\cite{ding2023enhancing} by HuggingFace, which consists of about 200k high-quality multi-round dialogues generated by ChatGPT that span a wide range of topics. For supervised fine-tune, the full dataset is used as training data while for SPIN and LD-Align, only a subset containing 50k samples is used.
\subsubsection{Model}
The model we used for our experiments, Mistral-7B\cite{jiang2023mistral}, is a widely used pre-trained LLM engineered for superior performance and efficiency. Mistral 7B outperforms Llama 2 13B across all evaluated benchmarks, and Llama 1 34B in reasoning, mathematics, and code generation. It should be noticed that, Huggingface has provided zephyr-7b-sft-full, which is fine-tuned on Ultrachat200k with Mistral-7B. We choose zephyr-7b-sft-full as our fine-tuned baseline and experiments conducted with selected competing methods were all based on this fine-tuned model.
\subsubsection{Evaluation metric}
We evaluate the performances of different methods using several benchmarks and reward scores.
For benckmark evaluation, we widely select datasets and benchmarks for evaluating the performance of models:
\begin{itemize}
    \item \textbf{Truthfulness.} TruthfulQA\cite{lin2021truthfulqa}
    \item \textbf{Commonsense Reasoning.} PIQA\cite{bisk2020piqa}, Hellaswag\cite{zellers2019hellaswag}, ARC\cite{clark2018think}, OpenbookQA\cite{mihaylov2018can}
    \item \textbf{CoT Reasoning.} GSM8k(CoT)\cite{cobbe2021training}
    \item \textbf{Multi-round Dialogue.} Mutual\cite{cui2020mutual}
\end{itemize}

For reward evaluation, our goal is to assess how well the model’s
response aligns with human preferences. We choose a widely used reward model based on DeBERTa\cite{he2020deberta} (reward-model-deberta-v3-large-v2) as the reward model. Given an instruction \textbf{x} and a response \textbf{y}, we leverage
a reward model RM to compute the reward score \textbf{r} = RM(\textbf{x}, \textbf{y}), which reflects how well the response \textbf{y} aligns with human preference with given instruction \textbf{x}.
\subsubsection{Competing Methods}
To show the efficiency of proposed LD-Align, we will compare LD-Align with several methods. We describe several most competing models as follows.

\textbf{Continual SFT} We continually train the model in a SFT fashion on Ultrachat-200k dataset. We set the lr to be 1e-5 and train for 2 epochs with a global batch size of 64 on multiple GPUs.

\textbf{SPIN}
SPIN is also a annotation-free alignment training method which try
to convert a weak model to a strong one through the lens of self-play, where the model is enhanced by playing against itself without requiring any direct supervision. We report the results of SPIN reproduced in our machine with the codes provided in their repository and we keep all the hyperparameters the same with 
 the setting mentioned in SPIN's paper.

\textbf{DPO}
DPO requires annotated prefernence data for training, we utilze UltraFeedback Binarized, which comprises both chosen and rejected chat completions evaluated by GPT-4. We train the zephyr-7b-sft-full with DPO  We set the lr to be 2e-7 and beta to be 0.1. We train the model for 2 epochs (which is already near to convergence) with a global batch size of 64 on multiple GPUs.
\subsection{Results}
It can be observed that following three iterations of LD-Aligh training, the model exhibited a notable improvement across multiple benchmarks. In the final iteration, the overall average enhancement reached 6.00\%, surpassing the performance of SPIN.

To validate the quality of the latent space characterized by the guiding model, we conducted the following analyses: (1) Analyzing the distribution of latent distances between sample pairs; (2) For a given input $\mathbf{x}$, we sampled and generated two responses $\mathbf{y^\prime_a},\mathbf{y^\prime_b}$ and calculate their  distances to real response $\mathbf{y}$. The response with the greater distance was considered lower in quality. Additionally, both responses were evaluated by GPT4 to judge which one is more aligned with real response, and we recorded the number of instances where their judgments are consistent.
\subsection{Conclusion}
We propose LD-Align, which can align model on the high-quality SFT datasets by leveraging the distances of samples in latent space without any additional manual annotations. Our approach yields significant improvements across multiple benchmarks, effectively enhancing the overall competence of the model

\bibliography{custom}
\bibliographystyle{acl_natbib}
\end{document}